\documentclass[table,times,11pt,a4paper]{article}

\usepackage{style/emnlp2020}

\newif\ifcomment\commenttrue

\aclfinalcopy

\usepackage{amsmath,amsfonts,bm}

\ifcomment
\newcommand{\pinaforecomment}[3]{\colorbox{#1}{\parbox{.8\linewidth}{#2: #3}}}
\else
\newcommand{\pinaforecomment}[3]{}
\fi

\def\eqref#1{equation~\ref{#1}}
\def\1{\bm{1}}

\def\evh{{h}}

\DeclareMathAlphabet{\mathsfit}{\encodingdefault}{\sfdefault}{m}{sl}
\SetMathAlphabet{\mathsfit}{bold}{\encodingdefault}{\sfdefault}{bx}{n}

\newcommand{\E}{\mathbb{E}}

\newcommand{\ind}[1]{\mathds{1}\left[ #1 \right] }

\DeclareMathOperator*{\argmax}{arg\,max}

\usepackage{hyphsubst}
\HyphSubstLet{english}{usenglishmax}
\usepackage[english]{babel}

\usepackage{xcolor}
\usepackage{amsmath}
\usepackage{booktabs}
\usepackage{hyperref}
\usepackage{mdwlist}
\usepackage{dsfont}
\usepackage{times}
\usepackage{url}
\usepackage[pdftex]{graphicx}
\usepackage[]{algorithm2e}
\usepackage{multirow}
\usepackage{subcaption}
\SetKwInOut{Parameter}{Parameters}

\title{Meta Answering for Machine Reading}

\author{\normalsize
        Benjamin B\"orschinger\textsuperscript{1},
        Jordan Boyd-Graber\textsuperscript{1,4},
        Christian Buck\textsuperscript{1},
        Jannis Bulian\textsuperscript{1},  \\ 
        {\bf \normalsize
        Michelle Chen Huebscher\textsuperscript{1},
        Massimiliano Ciaramita\textsuperscript{1},
        Wojciech Gajewski\textsuperscript{1},} \\
        {\bf \normalsize
        Yannic Kilcher\textsuperscript{2}\thanks{\, Contribution partially while as an intern at Google.}\hphantom{,}, 
        Rodrigo Nogueira\textsuperscript{3}\thanks{\, Contribution while an intern at Google.}\hphantom{,}, 
        Lierni Sestorain Saralegui\textsuperscript{1}}\\
\normalsize{
  \textsuperscript{1}Google Research \quad
  \textsuperscript{2}ETH Z\"urich \quad
  \textsuperscript{3}New York University \quad
  \textsuperscript{4}University of Maryland}\\
\texttt{\small \{bboerschinger, jbg, cbuck, jbulian, michellechen, massi, wgaj, lierni\}@google.com} \\
\texttt{\small yannic.kilcher@inf.ethz.ch} \quad \texttt{\small rodrigonogueira@nyu.edu}}

\newcommand{\g}{\, | \,}

\newcommand{\bs}[1]{\boldsymbol{#1}}

\newcommand{\name}{meta-answerer}

\newcommand{\clstok}{\texttt{{[}CLS{]}}}

\newcommand{\verminusone}{\textsc{AnswerOnly}}
\newcommand{\verzero}{\textsc{Context}}
\newcommand{\verthree}{\textsc{RewriteQues}}

\newcommand{\fone}{F1}
\newcommand{\bert}{\textsc{bert}}

\newcommand{\history}{evidence}
\newcommand{\History}{Evidence}
\newcommand{\inputa}{q, t, a_{i}, h}
\newcommand{\inputh}{q, t, h}
\newcommand{\nbest}{$M$-best}

\begin{document}

\maketitle

\begin{abstract}
We investigate a framework for machine reading inspired by real-world information-seeking tasks. Similarly to scrolling through web search results, a \name{} inspects the candidate answers to a question provided by a machine reading QA system.  
We find that with just a small snippet of text around an answer, humans can outperform the underlying QA system.
Similarly, a simple machine \name{} outperforms the reader on the Natural Questions dataset with equally impoverished information. 
However, humans and machines seem to differ crucially, in terms of strategies and capacity to use contextual information. 
Thus, the task suggests a challenge to probe the understanding of context in NLU.
\end{abstract}
\section{Introduction}

Question Answering (QA) is a benchmark task in Natural Language
Understanding (NLU) that has evolved from a niche
evaluation~\cite{voorhees-00} to a yardstick for measuring human
vs. computer intelligence~\cite{ferruci-10}.
The dominant format for QA is \emph{machine
  reading}~\citep{rajpurkar-etal-2016-squad}: answering a question
either by extracting a span of text identifying the \emph{answer} from
a single textual \emph{context}, or abstaining~\cite{rajpurkar-18}.

Machine learning can answer many questions using machine reading, but the price of this setting is artifice, as both the data and the task
fail to represent an ideal question answering setup.
On the data side, systems can shirk
understanding language in favor of artifact-enabled shortcuts, thus
making them vulnerable to adversarial
attacks~\citep{jia-liang-2017-adversarial,mudrakarta-etal-2018-model,niven-kao-2019-probing}.
On the task side, real-world information-seeking problems such as web search, have different properties; e.g., they are mediated by
machines which provide relevant but incomplete and noisy
information to a \emph{human user}.
Computers and humans have different skills and different
resources~\citep{Xie:2002:PII:566014.566017,Marchionini:2006,joy-of-search}.
Humans can synthesize discrete pieces of information, create
long-range strategies, and understand nuanced language.
Computers, in contrast, have nigh infinite memory and can take advantage of
subtle patterns in their inputs~\cite{wallace-19}.

To put humans and computers on an even playing field, we re-purpose the machine reading task as \emph{meta answering} (Section~\ref{sec:meta}). A \name{}, when presented with the output of a question answering system, must decide which (if any) of the top candidate answers is correct.
This task is characterized by imperfect information, since the QA system predictions are noisy. Information is also asymmetric: the \name{} and QA system have different views of the data. At the same time, the task is backward-compatible with respect to its original evaluation setup, hence, results are comparable.

We study how humans and computers perform, in both qualitative and quantitative terms, with an existing machine reading (MR) dataset---Natural
Questions~\citep{nq}---which embeds real users' information seeking needs in a realistic information retrieval context.
Our comparison recapitulates the divergent skills of humans and machines.
Humans do better when provided with more information (Section~\ref{sec:human}) because they discern whether
answer candidates are responsive to the question, can solve ambiguous
references, and can spot irrelevant distractors that can vex brittle
QA systems.
In contrast, the machine \name{} is overall better at the task
(Section~\ref{sec:exp}), besting a BERT~\citep{devlin2018bert} reader by two F1 points.
It falters, however, when provided with 
richer contexts.

Thus, meta answering surfaces limitations of NLU and provides insights on different processing strategies: machines are risk-averse, while humans seem naturally inclined to explore. Section~\ref{sec:conc} discusses how synergies between them could further improve NLU in information-seeking tasks.
\section{From Question Answering to Meta-answering}
\label{sec:meta}

A \name{} takes as input the highest scoring answers of an extractive QA system for a given question. The \name{} then tries to improve on the underlying system's result by selecting an answer that possibly differs from the system's top-scoring one.
Formally, an extractive QA system maps a question-context pair $\langle q, d\rangle$
to a set of $M$ answer candidates and their scores, $\{a_{i}, \sigma_{i}\}_{i=1}^M$;
all $a_{i}$ are subspans of $d$.
In our experiments we use Natural Questions~\citep{nq}, where $q$ is a web search query, $d$ is the highest
ranking Wikipedia page returned by Google for $q$, and each answer candidate is a short
span from $d$.

In web search, a user sees less information, the search results only, compared to the search engine that uses the full document collection.
Similarly, a \name{} has access to strictly less information than the underlying QA system and 
lacks direct access to $d$. Its goal is to find the best answer $\hat{a}$ within QA's \nbest{} candidate list.
Again, this resembles a human confronted with a search results composed of links and tiny snippets. Analogously, the QA system returns `snippets' of $d$ centered on the answer
candidates: a window of $K$ tokens to either side of the answer context.
For example, given the question ``who did the 49ers beat in the super bowls'', the \name{} must
decide whether the candidate answer "San Diego Chargers" is the right answer, given the context ``The 49ers steamrolled the \underline{San Diego Chargers} 49 -- 26 , at''.
One of the issues we aim to evaluate is the importance of this additional context for humans and machines.

%In the next section we show that humans can successfully deal with the meta-answering task.
\begin{table*}[t]
    \centering
        \begin{tabular}{lrrrrrrrrr}
        \toprule
         & \multicolumn{3}{c}{\verminusone{}}& \multicolumn{3}{c}{\verzero{}} & \multicolumn{3}{c}{\verthree{}} \\
         & Pr. & Rec. & F1 & Pr. & Rec. & F1& Pr. & Rec. & F1 \\
         \midrule
         NQ annotator & {\bf 57.9}& 46.4& {\bf 51.5}& {\bf 64.2} & 51.6 & {\bf 57.2} & 57.7 &	45.5 & 50.9\\
         \bert{}~\citep{bert-baseline} &56.4& 45.7 & 50.5& 60.1 & 47.2 & 52.9 & {\bf 67.2}& 56.4 & 61.4 \\
         avg. Human & 40.7 & 41.1 & 40.6& 48.9 & 50.9 & 49.8 & 54.5 & 56.7 & 55.5\\
         best Human & 39.4 & {\bf 48.8} & 43.6& 52.7 & {\bf 59.1} & 55.7& 60.1	& {\bf 66.8} &{\bf 63.3}\\
         $\hookrightarrow$ vs. NQ annotator & -18.5 & 2.4 & -7.9& -11.5 & 7.5 & -1.5 & 2.4 & 21.3 & 12.4 \\
         $\hookrightarrow$ vs. \texttt{BERT} & -17.1 & 3.1 & -6.9& -7.5&11.9&2.8&-7.1	&10.4&	1.9 \\
         \bottomrule
        \end{tabular}
    \caption{A boostrap evaluation to fairly compare original NQ annotators, human \name{}s, and the baseline QA system.  In the context with the richest interaction, the best human \name{} can improve over both the average NQ annotator and the baseline QA system, mostly through improving recall.}
    \label{table:string_match_results}
\end{table*}

\begin{table}[t]
    \centering
    \rowcolors{2}{gray!25}{white}
    \begin{tabular}{p{2cm}cc}
         & Answerable & Unanswerable \\
         \toprule
        {\bf Answer Correctly} & \textit{Right} & N/A \\
        {\bf Answer Wrong} & Neg & Fool \\
        {\bf No Answer} & Dead & \textit{Abstain} \\
        \bottomrule
    \end{tabular}
    \caption{Ontology for \name{} outcomes based on whether a question is answerable.  Right and Abstain are ``good'' outcomes for \name{}, while the other outcomes are incorrect.  Humans often provide answers when they should not.}
    \label{tab:ontology}
\end{table}

\section{Humans as \name{}s}
\label{sec:human}

To better understand the task and provide a benchmark, we place humans in the role of the \name{}.
Like the machine \name{} in the next section, they see the top candidates from a strong \bert{}-based~\cite{alberti-etal-2019-synthetic} system's \nbest{} list and must select which (if any) to provide as the final answer. On one hand, humans have world knowledge that computers do not. 
However, they are restricted to the same context as computers; they may instead be burdened by their innate knowledge rather than aided.

This section examines how well human \name{}s can find correct answers in settings with increasing complexity and information; a subset of these settings also correspond to those of our machine \name{}. 

{\bf \verminusone{}} shows only the question and candidate answers \emph{without context}; the \name{} needs to decide whether \underline{Central Germany} is a reasonable answer to ``what culture region is Germany a part of''.
{\bf \verzero{}} adds surrounding context: is ``Charles Osgood as the Narrator \underline{Jesse McCartney} as JoJo, the Mayor's son'' a good answer to ``who is JoJo in \textit{Horton Hears a Who}''.
This is identical to the information the machine \name{} uses in the next section.
{\bf \verthree{}} goes beyond the machine \name{}: users can ask questions to two other QA systems:  \citet{lee-etal-2019-latent} over all of Wikipedia or ask the system that generated the \nbest{} list a \emph{different} question over the NQ source page.
For example, the user asks ``Who did Jesse McCartney play in \textit{Horton Hears a Who}'' to verify (the system believes) he plays JoJo.

\subsection{Human Answering Framework}

A human \name{} interacts with the underlying QA system through a text-based interface (Figure~\ref{fig:interface} in Appendix).
They first see a prompt; they can then request an answer from the underlying QA system.
After requesting up to 20 examples, the user can either abstain or select one of the answer candidates.
For each condition, the same five human \name{}s (results are averages with error bars) play episodes for random samples of 100 questions from NQ dev questions.
In addition, in \verthree{}, the human \name{} can ask a different question as an action.

\subsection{Comparing Human \name{}s to NQ Annotators and the QA system}
\label{sec:human-vs-machine}

First, however, we need to discuss how to compare human \name{}s to a traditional QA system or to the humans who created the dataset. 
NQ uses \emph{exact span} as an error metric,\footnote{Both human and computer \name{}s also cannot answer questions with disjoint spans; e.g., names from a cast or
binary yes/no questions.} however this does not work for human \name{}s: our interface shows a span out of context so humans do not know where the string came from.
Thus, selecting the correct answer at the wrong position is
still counted as a miss (e.g., \underline{Kevin Kline} at token 13 or 47 is wrong but \underline{Kevin Kline} at token 30 is correct). 
To compare the accuracy of the QA system and original NQ annotators (both see the whole source document) with the humans operating on partial knowledge, we
compare both \emph{exact string} matches and partial string matches (\emph{surface F1} measuring token overlap).

The first question is whether our human \name{} are as good as the original NQ annotators (who see the whole document).
NQ provides five annotations of potential answers; and answer ``counts'' if two NQ annotators agree.
For a fair apples to apples comparison, we would need to hold out one NQ annotation and compare it to our human \name{}s.

Rather than rely on a deficient gold set, bootstrap sample  
from the remaining four annotations to get back to five annotations.
But it would unfair to only evaluate NQ annotators using this, hence we evaluate our humans and the baseline \bert{} QA system using the same metric (a bootstrap sample to get five annotations).
This is a fair comparison between the NQ annotator, our
human annotators, and the baseline QA system. 
Human \name{}s have lower F1 than both the original annotators as well as the baseline QA system (Table~\ref{table:string_match_results}), in \verzero{} and especially  \verminusone{}.
All humans improve between \verminusone{} and \verzero{}.
There is also an improvement from \verzero{} to \verthree{}, but formulating new questions also introduces more variance,\footnote{The highest agreement between raters is near 0.7 chance-adjusted $\kappa$, while agreement between the users and the system is typically between 0.5 and 0.6.} which is a function of skill.  
The better human \name{}s improve over the baseline system.
The best human (consistent across all conditions) was $-7.9$ worse on \fone{} in the \verminusone{}, the best human \name{} has a higher \fone{} than the average NQ annotator in \verthree{} with substantially less information.

\subsection{A Taxonomy of Outcomes}
\begin{figure}[t]
     \centering
     \includegraphics[width=1.0\linewidth]{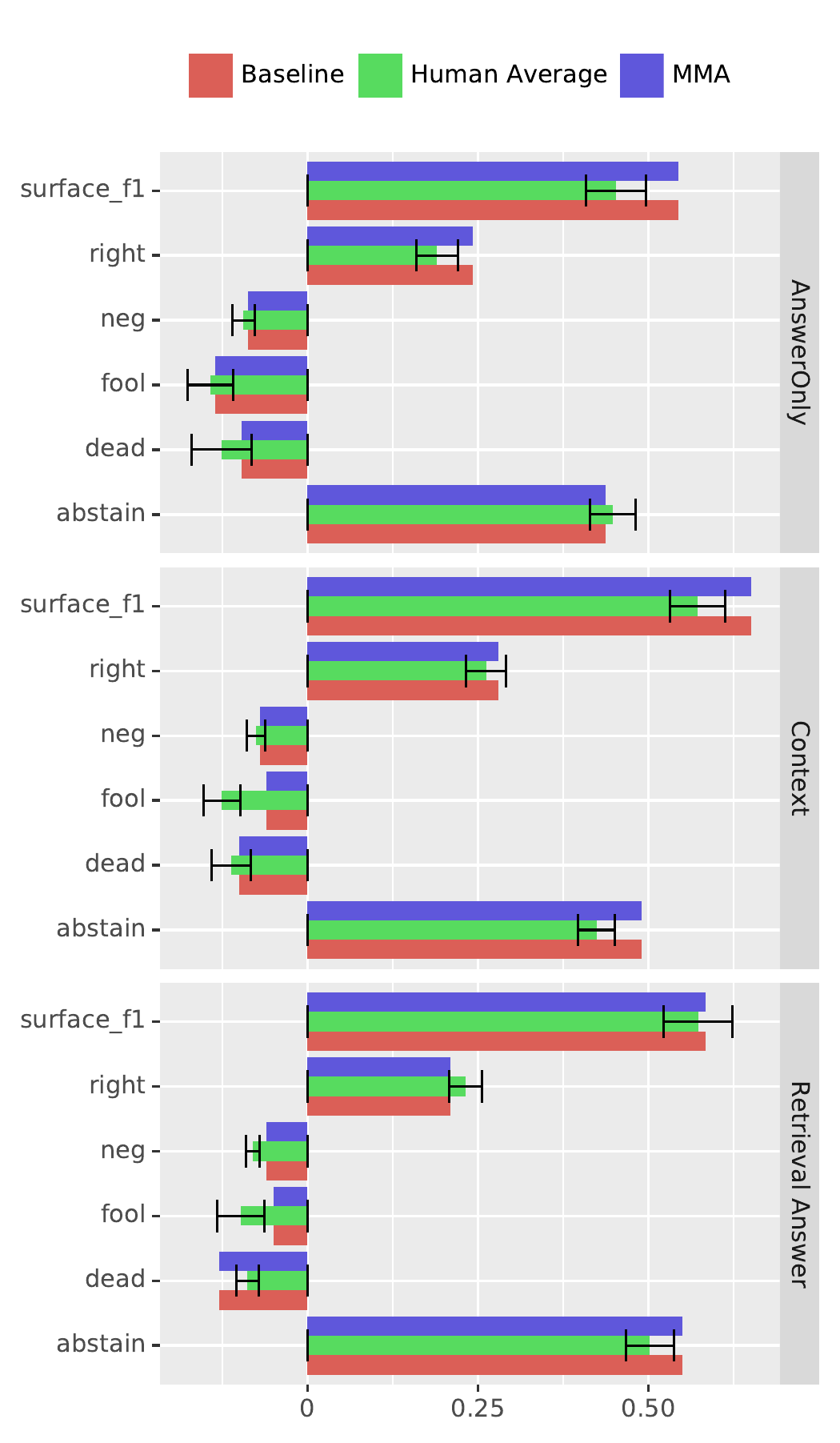}
     \caption{Humans improve over the underlying QA as they see more information.      they abstain less,      but this is often balanced out by being ``fooled'' (tricked into answering questions where they should abstain).  Numbers not comparable to bootstrap evaluation in Table~\ref{table:string_match_results}.}
     \label{fig:improve_over_qa}
\end{figure}

The human \name{}s improve recall far more than precision.  
To better understand where these gains come from, it's helpful to go beyond these broad outcomes.  
To see why human \name{}s have better recall, we describe possible outcomes of a \name{} (Table~\ref{tab:ontology}).  

\paragraph{\bf Right and Neg} 
The clearest result is that the \name{} selects a \emph{right} answer from the QA system and provides it.  
This can either be confirming the answer that the QA system would have presented, answering when the QA system would have abstained, or selecting a different answer.
The most human \name{} improvement  is to answer instead of abstaining: \underline{1967} to the question ``when did colour tv come out in uk'', the answer is at the top of the baseline QA's $n$-best list, but below the threshold.

Sometimes humans select a wrong answer when an answer is available.  
We call this a negative selection, or ``neg'' for short.
For example, the question ``when did the crucifix become the symbol of christianity'' has gold answers \underline{the 4th century}, \underline{in the 2nd century}, and \underline{4th century}.
A human \name{} selects \underline{the 2nd century}, which was not an acceptable answer.

\paragraph{\bf Abstain}
Many NQ instances cannot be answered\footnote{E.g., ``universal social services are provided to members of society based on their income or means'' is not a question, the NQ source provides an article about Paralympics for an Olympics question, or no Wikipedia page answers ``where am i on the steelers waiting list''.} the next most common outcome is for a \name{} to correctly recognize it should not provide a response.
The QA system abstains more than humans, which leads to human \name{}s biggest failing\dots

\paragraph{\bf Fool}
The flipside is being ``fooled'' into answering instead of abstaining; for example answering \underline{3 September} to the question ``when did the united kingdom entered world war 2'': humans are enticed by the context ``after a British ultimatum''.
While humans have a much higher rate of being fooled than computers (humans are reluctant to say ``I don't know''), some of this is attributable to annotation problems with NQ.

\paragraph{\bf Dead}
Sometimes a question is answerable, but the \name{} falsely abstains.
We call this result a ``dead'' question.
Sometimes the \name{} is unsure of an answer.  For example, for the question ``who was fighting in the civil war in england and about what'', the gold answer is ``Parliamentarians and Royalists over, principally, the manner of England's government''.
However, humans rejected that answer (perhaps as being too superficial).

Humans improve as we move to the \verthree{} setting, adding the ability to ask new questions.
Human \name{}s can improve recall by converting baseline abstentions into right answers (and a smaller number of baseline negs). However, humans are more often fooled, resulting in lower average precision (Figure~\ref{fig:improve_over_qa}).
Because the human \name{} is at the mercy of the baseline QA system, if the baseline system does not surface the answer, the question will go dead without the ability of the human to find the answer.

\subsection{Human Strategies}

Without contexts in the \verminusone{} setting, humans are limited to examining the question answer combination (e.g., knowing that ``Germany'' is not a part of \underline{Central Germany} or that \underline{Jennifer} is not the ``meaning of the name Sinead'').
However, these cases are rare enough that human \name{}s do not overall improve the system.

With \verzero{}, human \name{}s use context to select better answers than the system.  
For example, seeing that \underline{near Arenosa Creek and Matagorda Bay} was settled by explorer Robert Cavelier de La Salle in the Wikipedia page context, allowing to convert a dead model question into a correct one.

With \verthree{}, humans can more thoroughly probe the source document to establish whether an answer is correct. 
E.g., while the baseline system answers the question ``who plays eddie's father on blue bloods'' with \underline{Eric Laneuville}, humans can explore outside the source document to find the correct answer (William Sadler, who plays Armin Janko, only appears in the sixth season, while the NQ document is about an earlier season) or within the page to establish that Eric Laneuville directed several episodes of \textit{Blue Bloods}.

\begin{table*}[t]
    \centering
        \includegraphics[width=1.0\linewidth]{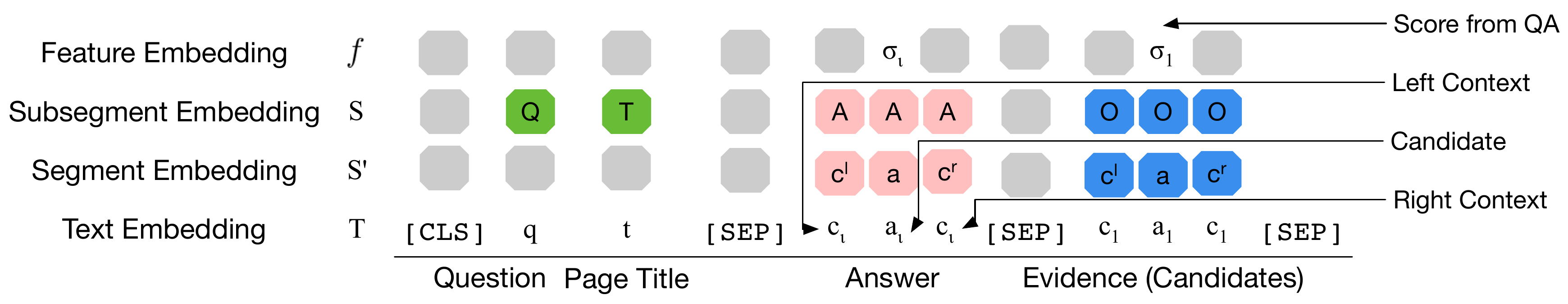}
    \caption{Our \name{} has three primary components to determine answers, \history{} sequences, and impossibility.  Each uses a representation from a BERT representation that distinguishes (Segment embedding) the Wikipedia page that contains the answer, the answer candidate, and \history{} that encodes alternative candidates as well as the internal structure of the answer and \history{} (Segment Embedding).  The feature embedding layer captures the strength of the original QA model's prediction.}    \label{table:state_representation}
\end{table*}

\section{Machines as \name{}}
\label{sec:machine}

Like the humans, machine \name{}s inspect and evaluate candidates, based on the available context, and then provide an answer (or abstain).
The most salient difference from traditional machine reading systems is that the \name{} does not have access to the entire text---it must choose from discrete pieces of \history{} to justify which (if any) answer to return.
We break the task into three subtasks with their own losses: an answer candidate selector~$P_A$, an \history{} selector~$P_H$ to decide which evidence to pick,
and an impossibility classifier~$P_I$ to decide whether to answer at all.

The {\bf \history{}} are the candidate answers and the surrounding context around the answer (as with the human \name{} in Section~\ref{sec:human}).
While humans can skim over other answers and their contexts to use salient hints, we force the computer to explicitly build a set of \history{}.

\subsection{System architecture}

Our classifiers are output layers on top of  a BERT encoder~\citep{devlin2018bert} which translates a semi-structured input into a  single dense vector.

The $P_A(\cdot \mid \inputa{})$ is 
the probability that candidate~$a$ is a correct answer to question~$q$,  assuming that the (unobserved) 
Wikipedia page from which $a$ is extracted has title~$t$, and an \history{} set of hints~$h = \langle o_{1}, \dots, o_{K} \rangle$;
these inputs are encoded into a dense vector using a BERT model parameterized via $\bs{W}_E$
\begin{equation}
    \bs{e}_{A}^{(\clstok)}   = \textsc{BERT}_{\bs{W}_E}(\bs{E}(\inputa{}))
    \label{eq:inputa}
\end{equation}
and then becomes a probability as the output of the feed-forward neural network (FFNN):
\begin{equation}
    P_{A}(\cdot \mid \inputa{}) = \textsc{FFNN}_{\bs{W}_A}(\bs{e}_{A}^{(\clstok)}).
    \label{eq:pa}
\end{equation}

The individual hints in the \history{} set are chosen by $P_H$ as `relevant' for answering through a similar process.
The inputs to evaluate a piece of \history{} are the inputs to evaluate a candidate except there is no answer~$a_i$, as the goal is for the \history{} set to be useful for \emph{all} candidates.
Given this input of question, title, and set of \history{}~$h$, we create a dense representation
\begin{equation}
    \bs{e}_{H}^{(\clstok)}   = \textsc{BERT}(\bs{E}(\inputh{})),
    \label{eq:inputb}
\end{equation}
which, as above, is turned into a probability via a feed-forward network,
\begin{equation}
    P_{H}(\cdot \mid \inputh{})  = \textsc{FFNN}_{\bs{W}_H}(\bs{e}_{H}^{(\clstok)}).
\end{equation}

\subsection{Semi-structured Embeddings}
\label{sect:sse}

An effective input representation for \bert{} must encode not just individual words but also the role those words play.
In Equation~\ref{eq:inputa} and~\ref{eq:inputb}, we differentiate the words in the context around an answer and the title of a source page.
In addition to an embedding matrix for text tokens~$\bs{E^{T}}$, we use a top-level segment embedding to distinguish questions~$Q$, title~$T$, answer candidates~$A$, and \history{}~$O$ (row $S$ in Table~\ref{table:state_representation}).

There is a finer distinction within the answer candidates and \history{}.
We need to separate the context around an answer candidate and the candidate itself (e.g., distinguishing the token ``Nigel'' from its context in ``he voiced  \underline{Nigel} in Rio'').
We use sub-segment embedding~$\bs{E^{S'}}$ to distinguish the answer span~$a$ and its surrounding left ($c^{(l)}$) and right ($c^{(r)}$) context, within each primary segment type (either Answer~$A$ or \History{}~$O$).
To the text inputs we concatenate the score of the candidate~$\sigma_i$ (row $f$ in Table~\ref{table:state_representation}) and finally the position embeddings $\bs{E^{P}}$:
\begin{equation}
    \bs{E}_{:,i} = (\bs{E^{T}}_{:,T_{i}}, \bs{E^{S}}_{:,S_{i}}, 
                    \bs{E^{S'}}_{:, S'_{i}}, F_{i}\times \bs{e^{f}}, \bs{E^{P}}_{:,i}).
\end{equation}
When encoding inputs that should not contain the answers (see below) we mask the respective tokens.

\subsection{Answer Candidate Selector}
\label{sec:training}

We train $P_{A}(\cdot \mid \inputa{})$ with a binary cross-entropy loss, assuming examples of the form
$\langle y, q, t, o, h\rangle$, where $q$ is a question, $t$ a page title, $o=(c_i^{(l)},a_i,c_i^{(r)})$ a candidate
answer in context, $h$ the \history{}, and a binary label~$y$ indicating whether the answer is a correct answer.
We train this classifier using the \nbest{} list from \citet{bert-baseline}.
Because most candidates are wrong, we downsample negative examples.\footnote{Different negatives are picked at each epoch so that in expectation all negative examples are seen.} 

Selecting the \history{} sequence is less straightforward; we describe this in the next section. 
For the moment, however, let us assume that we have an \history{} sequence~$h$ for each question.
Then for a $\mathbb{D}$, the answer selector loss $\mathcal{L}_{A,\mathbb{D}}(\bs{W}_{A}, \bs{W}_{E})$ is
\begin{equation}
\sum_{\langle y, q, t, o, h\rangle \in \mathbb{D}} -\log P_{A}(y \mid \inputa{}; \bs{W}_{A}, \bs{W}_{E}). \notag
\end{equation}

\subsection{\History{} Selector}

The \history{} selector uses $P_{H}(\,\cdot \mid \inputh{})$---the probability that $h$ is a useful sequence
of observations to evaluate answers for question~$q$ and title~$t$ using $P_{A}$---to pick, among a set 
of candidate \history{}s $\mathcal{H}$, the sequence~$h$ of length $M$ that scores the highest. 

Training $P_H$ requires knowing the answer because we want it to score good \history{} sequences higher.
But we don't want $P_H$ to directly depend on the answer.
To create an implicit training signal, we induce a pseudo-label that does depend on the answer that is 1 whenever $h'$ provides better \history{} for determining the correct answer $d(P_A, q, t, o, y, h, h') \equiv$
\begin{equation}
\label{eq:pseudo-label}
\ind{P_{A}(y\mid q, t, o, h)  < P_{A}(y\mid q, t, o, h') } 
\end{equation}
For readability, we abbreviate Equation~\ref{eq:pseudo-label} as $d(P_A,h,h')$.

Given this pseudo-label for training, we train $P_H$ with the answer candidate~$a_{i}$ masked (Table~\ref{table:state_representation}) to allow $P_H$ to discover useful properties of the \history{} to find the correct answer.
For
a pair of candidate evidences $h$ and $h'$, this induces a \history{} loss $\mathcal{L}_{H,\mathbb{D}}(\bs{W}_{H}, \bs{W}_{E})$ to train $P_H$,
\begin{align*}
& \sum_{\langle y, q, t, o, h\rangle \in \mathbb{D},h'}\!\!\!\!\!\!\!\!\!\!\!\left[-\log P_{H}(d(P_A,h,h')\mid q, t, h; \bs{W}_{H}, \bs{W}_{E}) \right. \notag \\
& \hphantom{\dots} \left. -\log P_{H}(d(P_A,h',h)\mid q, t, h'; \bs{W}_{H}, \bs{W}_{E})\right].
\end{align*}
We generate an alternate \history{}~$h'$ by randomly replacing one of the observations in $h$.\footnote{Decoding also selects $h, h'$ with unit Hamming distance.}

The \history{} loss thus maximizes the expected reduction in entropy for $P_A$ and $q$ produced by substituting $h$ with $h'$.
While the full expectation would require summing over all possible answers for $q$, training on a single answer defines an unbiased (although noisy) estimate of this expectation.
The loss and, consequently, the training signal for $P_H$ implicitly depends on
a $P_A$. 
However, $P_A$ and $P_H$ can be co-trained from scratch.

\subsection{Auxiliary impossibility loss}

51\% of the questions in the NQ dataset are `unanswerable', i.e., there is no gold answer in the context. As demonstrated in the human \name{} experiments, learning to abstain is crucial for doing well on NQ.

To help our model detect unanswerable questions, we add an additional impossibility classifier~$P_I$.
The impossibility classifier is a single feed-forward layer on top of $\bs{e}^{\clstok}_A$ (Equation~\ref{eq:inputa}) with loss $\mathcal{L}_{I,\mathbb{D'}}(\bs{W}_{I}, \bs{W}_{E}) =$
\begin{equation}
    \sum_{\langle q, t, b, o, h\rangle \in \mathbb{D'}} -\log P_{I}(b\mid q, t, o, h; \Theta_{I}, \Theta_{E}),
\end{equation}
where $b$ is a binary label that is $1$ for answerable examples and $0$ otherwise.

\subsection{Training}

We initialize \bert{} from the public checkpoint. Because \bert{} is most effective with masked language model pre-training (MLM),
we ensure our encoder is used to the input described in Table~\ref{table:state_representation} using examples
in $\mathbb{D}$ to compute input sequences and randomly masking 30 tokens in each.  We run the pre-training
for 200,000 steps, using a batch size of 32.

Our pre-training produces better models, as does cotraining $P_A$, $P_H$, $P_I$ and MLM: masking one random token from every input.
We combine the four losses into a single weighted loss and treat the per-loss weights
as hyper-parameters:
\begin{equation*}
\mathcal{L} = \lambda_{A}\mathcal{L}_{A} + \lambda_{H}\mathcal{L}_{H} + \lambda_{I}\mathcal{L}_{I} + \lambda_{MLM}\mathcal{L}_{MLM}
\end{equation*}

\paragraph{Making predictions}

\begin{algorithm}[t]
\SetKwFunction{GetCandidates}{GetCandidates}
\SetKwFunction{ComputeAnswerScores}{ComputeAnswerScores}
\SetKwFunction{MaxScoreAnswer}{MaxScoreAnswer}
\SetAlgoLined
\KwIn{$q, t, \mathrm{QA}, M$}
\Parameter{$k,P_A,P_H$}
\KwOut{$\vec{s}_{a_i}$}
$\mathcal{A} \leftarrow$ \GetCandidates{$q,t, \mathrm{QA}, M$}\;
 $\displaystyle \evh_{1:k}\leftarrow \mathcal{A}_{1:k}$\;
  \For{$t=k+1$ \KwTo $M$}{
$i \leftarrow \argmax_{i\in \{1,\dots,k\}} P_H(\mathrm{good} \g q, t, h_{i/t})$\;
 \If{$P_H(\mathrm{good} \g q, t, h_{i/t})>P_H(\mathrm{good} \g q, t, h)$}{
    $h_i\leftarrow h_t$\;
 }
 }
\For{$i=1$ \KwTo $M$}{
 $\vec{s}_{a_i}[i] \leftarrow P_A(\mathrm{correct} \g q, t, a_i, h) $\\
 }
 \caption{Scoring algorithm for the \name{} paired with MR system QA.}
  \label{alg:greedy}
 
\end{algorithm}

We now describe
how, at test time, we generate answer predictions.
Algorithm~\ref{alg:greedy} uses $P_A$ and $P_H$ to serve as a \name{} on top of an existing QA system.
It first builds a size-$k$ \history{} set from the top-$k$ answer candidates
from the QA system. It then iterates over the remaining $M-k$ candidates using $P_H$ to greedily decide whether to replace observations in the \history{}. For example, $h_{i/j}$ is the \history{} when the $i^{th}$ element of $h$ has been replaced with observation $o_j$.
Once all $M$ candidates have been processed and a $k$-size history $h$ has been selected, we score
all candidates using $P_{A}$ and return their scores.
We do not use $P_I$ at test time.\footnote{Nevertheless, co-training $P_I$ is beneficial as this auxiliary training loss helps create representations that better capture whether a question is impossible.}
Following \citet{alberti-etal-2019-synthetic}, we always predict the best answer if the score is above a threshold picked on development data.

\begin{table*}[t!]
\begin{center}
\begin{tabular}{lcccccc}
\toprule
\multirow{2}{*}{\hphantom{\dots\dots\dots\dots\dots\dots} System} &\multicolumn{3}{c}{Short Answer Dev}&\multicolumn{3}{c}{Short Answer Test}\\
&P&R&Span F1&P&R&Span F1\\\midrule
Single annotator~\citep{nq} & 63.4 & 52.6 & 57.5 & - & - & - \\
\texttt{BERT}~\citep{bert-baseline}& 59.5 & 47.3 & 52.7 & 63.8 & 44.0 & 52.1\\
\texttt{BERT\textsubscript{WWM}} & 59.5 & 51.9 & 55.4 & 63.1 & 51.4 & 56.6 \\
\texttt{MMA\textsubscript{Base}} &56.7&49.5&52.9&55.5&49.4&52.3\\
\texttt{MMA}$_{\verminusone{}}$ & 63.1 & \textbf{54.3}  & 58.4 & \textbf{67.3} & 52.4 & \textbf{58.9} \\
\texttt{MMA}$_{\verzero{}}$ & \textbf{64.5} & 53.5 & \textbf{58.5} & 66.2 & \textbf{52.7} & 58.7 \\
\bottomrule
\end{tabular}
\end{center}
\caption{Results on the Short Answer task of the Natural Questions dataset.}\label{tab:results}
\end{table*}

\section{Machine Experiments}
\label{sec:exp}

\subsection{Machine Meta Answering systems}
We evaluate three versions of the machine \name{}s.
The first, \texttt{MMA\textsubscript{Base}}, is a baseline to see if a model can pick out the correct answer without context and minimal textual evidence. For each $a_i$ in the \nbest{} list of \texttt{BERT\textsubscript{WWM}}, we encode an observation as $(a_i,q,a_1,a_2,\ldots,a_M)$. We annotate $a_i$ with segment id “A”, the rest with “B”. Answers and question are separated by [SEP] tokens. Overflowing text is truncated from the end. On the last \clstok vector we add a binary classifier, corresponding to $P_A$ in Equation~(\ref{eq:pa}), trained to identify the gold answer labels for the candidate $a_i$. \texttt{MMA\textsubscript{Base}} is a vanilla BERT implementation; i.e., without the semi-structured embeddings corresponding to the top two rows of Table~\ref{table:state_representation}, without evidence selection and multiple losses described in Sections~(4.1-4.6).

We then focus on two full versions of the machine \name{}s, \texttt{MMA}$_{\verminusone{}{}}$ and  \texttt{MMA}$_{\verzero{}}$. As in the human experiments, the former can only see the text of the candidate answer, while the latter can also see a snippet of five tokens to each side of the answer.
All losses are trained jointly for 200,000 steps, initializing the encoder with our
custom pre-trained BERT large model and randomly initializing $\bs{W}_{A}, \bs{W}_{I}$ and $\bs{W}_{H}$.
After hyper-parameter search, for \texttt{MMA}$_{\verminusone{}{}}$ we set the $M$-best list size $M=5$, evidence size $k=3$, and $w_{A}=3, w_{H}=0.1, w_{I}=10, w_{MLM}=0$, for the weighted training loss. For \texttt{MMA}$_{\verzero{}}$ we set $M=4$, $k=3$, $w_{A}=3, w_{H}=10, w_{I}=3, w_{MLM}=1$.

\subsection{Results}
Full results on the NQ short answer task are reported in Table~\ref{tab:results}.
We take as reference \citet{bert-baseline}'s QA model trained on NQ and
refer to it as \texttt{BERT}.  Adding whole word masking~\citep{roberta_paper} adds
 4.5 \fone{} points, and we use this \texttt{BERT\textsubscript{WWM}} as the 
QA system. The baseline, \texttt{MMA\textsubscript{Base}}, is worse than \texttt{BERT\textsubscript{WWM}}. This shows that a na\"ive approach to encode evidential information is not comparable with the original answer context. The more careful handling of evidence, embeddings, and multiple losses lead the full \texttt{MMA} systems to outperforming \texttt{MMA\textsubscript{Base}} by more than 6 F1 points on test. They also improve over \texttt{BERT\textsubscript{WWM}} by about 2 F1 points. Interestingly, \texttt{MMA}$_{\verminusone{}}$ outperforms \texttt{MMA}$_{\verzero{}}$ on test, particularly on precision (see Appendix~\ref{sec:analysis} for breakdown), while the latter has slightly better recall.
This is counter intuitive, from a human perspective, as humans are better with more information. 
Together with the additional finding that this is flipped on dev---\texttt{MMA}$_{\verzero}$ has higher precision and F1 than \texttt{MMA}$_{\verminusone}$---indicates that more work is needed to properly model context.

\subsection{Comparing Computers and Humans}

In both cases, the \name{} works with a stream of answer candidates and tries to improve the top prediction.  
The computer \name{} slightly improves both precision and recall. Qualitatively, the changes are minor: tweaking a span, adding a word, favoring earlier spans, improving the abstention threshold.

Human \name{}s, however, were more bold.  This allows them to greatly improve recall, digging deep in the \nbest{} list to find the answers they believe best answer the question.  This exploratory behavior comes at the cost of lower precision; they often get fooled by plausible sounding answers that the NQ annotators did not agree with.

 Interestingly, while humans do better when given more context, machine \name{}s did not.   We suspect that the models are still not rich enough for the computer to reason with using the augmented evidence provided in the contexts.

\section{Related Work}
\label{sec:related}

The work carried out in the early 2000s in the context of the TREC conference,\footnote{\url{https://trec.nist.gov/}.}, QA track~\cite{voorhees-00} is relevant here. The evaluation of the evidence (candidate answers and their context) is an important aspect of meta answering. The TREC QA framework envisions accounting for answer support as a key part of the task~\citep{voorhees-03} and successful systems relied crucially on notions of “answer justification”~\citep{Harabagiu00falcon:boosting}. A substantial difference\footnote{Besides the evolution of the machine learning paradigm, and  associated size requirements for datasets.} is that a \name{} has no way of affecting the production of evidence (the retrieval step) which is provided by the reader. Also relevant from this body of work, \citet{pizzato-molla-2005-extracting} propose a metasearch QA framework, which relies on several Web search and QA systems.
Another relevant line of research involves query reformulation, a popular method among search users~\citep{jansen2009}, mastered also by children~\citep{rutter-etal-2015}, which often relies on understanding and reusing the search results context~\citep{huang2009}. Automatic query reformulation has been applied with some success to machine reading. 
In \citet{Nogueira:2017} and \citet{aqa-iclr:2018}, reinforcement learning (RL) trained agents seek good answers while learning to reformulate questions.
\citet{das2018multistep} propose to perform query reformulation in embedding (continuous) space, and find that it can outperform the surface language reformulations of~\citet{aqa-iclr:2018}.
We do not consider reformulation here, although it may be a promising direction to explore. More generally, a \name{} differs from the methods discussed above in that it does not try to change the candidate list content.In this sense, it is akin to evaluating search results by scrolling down a static result list.

A practical solution to the limited encoding capacity of BERT is to summarize documents. \citet{han-etal-2019-episodic} propose an episodic reader that summarizes using RL. \citet{nishida-etal-2019-answering} combine answering and summarization to justify answers, which is effective when explanations are annotated~\citep{yang-etal-2018-hotpotqa,thorne-etal-2018-fact}. Yet, there is no evidence that summarization-based readers are competitive in major MR tasks.
Similarly related is skim-reading~\citep{seo2018neural, hansen2018neural}, where documents are only partially encoded for prediction. Skip actions are not supervised and systems are trained with RL. The main advantage is, again, efficiency.
\citet{yuan2019imrc} extend MR by re-purposing existing datasets such as SQuAD and NewsQA~\citep{trischler-etal-2017-newsqa} for a partially-observed, incremental, QA task. Here an agent is trained with RL for information-gathering and answering actions However, the performance of such agents is far from competitive.

%This is an important aspect  of language-based information-seeking tasks.
\section{Conclusion}
\label{sec:conc}

Meta-answering is a framework for QA that attempts to simulate real-world---imperfect---information-seeking tasks, where humans look for answers in settings mediated by machines, using natural language.
Human \name{}s can compete with a BERT-based single system with access to full documents, by only looking at a five token window around candidates. 
A machine \name{} built on BERT can improve the environment's QA system, thus proving that it is possible to investigate MR in imperfect information settings in high-performance regimes.
Further, the task brings to the surface, yet again but from a novel perspective, limitations of the current NLU paradigm.
MMA cannot use the contextual information that is effortlessly exploited by humans. Thus, it might prove a suitable framework to advance on these challenges.

For future work, we plan to research the use of reinforcement learning for meta answering. The motivation is two-fold. First, the sequential aspect of information-gathering and answering lend itself naturally to episodic formalization characterized by delayed rewards. Secondly, information seeking tasks might offer challenging and realistic problems for RL.\footnote{Preliminary experiments with RL to train a MMA with advantage actor critic~\citep{pmlr-v80-espeholt18a} failed to generalize beyond the trivial policy of copying the top prediction.} As rewards are sparse, credit assignment is hard. 

Language and cognition could help boost progress on more natural learning algorithms (cf.~\citet{Hung2019}), for example using imitation learning from human \name{}s to teach computers more intricate answering and verification strategies.  Together, this may allow computers to combine thoroughness with human flexibility and insight.

\bibliography{bib/iclr2020_conference,bib/journal-full,bib/jbg}
\bibliographystyle{style/acl_natbib}

\clearpage
\setcounter{section}{0}
\renewcommand{\thesection}{\Alph{section}} 

\section{Appendix}
\subsection{Differences between \texttt{MMA$_\verminusone$} and \texttt{BERT\textsubscript{WWM}}}
\label{sec:analysis}

Table~\ref{tab:analysis} breaks down the abstain and answer decisions for \texttt{MMA$_\verminusone$} and \texttt{BERT\textsubscript{WWM}}.
We find that \texttt{MMA$_\verminusone$} is a slightly more accurate abstainer, but that by and large, its tendency to answer
or abstain matches that of \texttt{BERT\textsubscript{WWM}}.  However, it is a considerably more accurate answerer -- even though
it answers $38$ times fewer than \texttt{BERT\textsubscript{WWM}}, it increases the absolute number of correct answers
given to by $83$ and decreases the number of incorrect answers given by $121$, explaining both the higher precision
and recall for the NQ evaluation metric.

\begin{table}[ht!]
\centering
\resizebox{\columnwidth}{!}{  
\begin{tabular}{|c|c|c|c|c|}
\hline 
\multirow{2}{*}{} & \multicolumn{2}{c|}{ABSTAIN} & \multicolumn{2}{c|}{ANSWER}\tabularnewline
\cline{2-5} \cline{3-5} \cline{4-5} \cline{5-5} 
& Correct & Incorrect & Correct & Incorrect\tabularnewline
\hline 
\multirow{3}{*}{\texttt{MMA}$_{\verminusone}$} & 3803 & 1050 & 1879 & 1098\tabularnewline
\cline{2-5} \cline{3-5} \cline{4-5} \cline{5-5} 
& \multicolumn{2}{c|}{78.36\%} & \multicolumn{2}{c|}{63.12\%}\tabularnewline
\cline{2-5} \cline{3-5} \cline{4-5} \cline{5-5} 
& \multicolumn{2}{c|}{4853} & \multicolumn{2}{c|}{2977}\tabularnewline
\hline 
\multirow{3}{*}{\texttt{BERT\textsubscript{WWM}}} & 3764 & 1051 & 1796 & 1219\tabularnewline
\cline{2-5} \cline{3-5} \cline{4-5} \cline{5-5} 
& \multicolumn{2}{c|}{78.17\%} & \multicolumn{2}{c|}{59.57\%}\tabularnewline
\cline{2-5} \cline{3-5} \cline{4-5} \cline{5-5} 
& \multicolumn{2}{c|}{4815} & \multicolumn{2}{c|}{3015}\tabularnewline
\hline 
\multirow{2}{*}{Difference} & 39 & -1 & 83 & -121\tabularnewline
\cline{2-5} \cline{3-5} \cline{4-5} \cline{5-5} 
& \multicolumn{2}{c|}{38} & \multicolumn{2}{c|}{-38}\tabularnewline
\hline
\end{tabular}
}
\caption{Overall accuracy for the original model and the agent, split by action type.}\label{tab:analysis}
\end{table}

\subsection{Qualitative analysis of differences between \texttt{MMA$_\verminusone$} and \texttt{BERT\textsubscript{WWM}}}

In this section, we will drill down into cases where the \texttt{MMA$_\verminusone$} answer has flipped from incorrect to correct, or vice-versa. Additionally, these cases are analyzed with respect to their nature, and grouped into three categories:
\begin{enumerate}
\item \emph{Entity change}: a chosen answer contains a different entity,
e.g. a different time-span, geographic name, or person. For the question:
\emph{When was the first time the Internet was used?} a correct answer: \textbf{1969} is chosen instead of the previously incorrect choice of \textbf{the 1980s}.
\item \emph{Different span/same entity}, e.g. for question: \emph{The vascular layer of the eye is the }A
different text span quoting \textbf{uvea} as an answer is chosen.
\item \emph{Incorrect span}: for a question the same entity in the same
text fragment is chosen, however, the original answer provides fewer or more details judged unnecessary for the answer by the raters, e.g. for the question:
\emph{Who is Tinker Air Force Base named after? }the span: \textbf{Oklahoma native Major General Clarence L. Tinker} is chosen instead of \textbf{Major General Clarence L. Tinker}.
\end{enumerate}

Statistics of the answer changes flips from incorrect to correct ("Loss to Win"), and from correct to incorrect ("Win to Loss") are listed in Table~\ref{tab:analysis_qualitative}.

\begin{table}[h]
\centering
\begin{tabular}{|c|c|c|}
\hline 
Modification &Loss to Win&Win to Loss\tabularnewline\hline
Entity Change &10 & 6\\
Different Span & 1 & 0 \\
Incorrect Span & 25 & 10\\\hline
\textbf{Overall} & 36 & 16\\
\hline
\end{tabular}
\caption{Classification of cases where the agent's answer changes correctness.}
\label{tab:analysis_qualitative}
\end{table}

Additionally, in case of Correct to Incorrect label change, we have found three answers to be incorrectly or doubtfully marked by the raters.

Moreover, among all of the cases (35 of them in total) where the algorithm has chosen to pick a different span with the original answer, in 31 of cases the span is shorter. That may indicate that the algorithm finds it more beneficial to avoid all of the unnecessary information from the answer, leading to an overall improvement for the answer statistics. In most of these cases, however, a human would probably judge both of the answers correct. For example, for question \emph{What class of ship is the Carnival Glory?} the original and deemed correct answer is \textbf{Conquest - class cruise ship} while the one picked by \texttt{MMA$_\verminusone$} - \textbf{Conquest - class} is labeled as incorrect by the raters.

\begin{figure*}

\includegraphics[width=\linewidth]{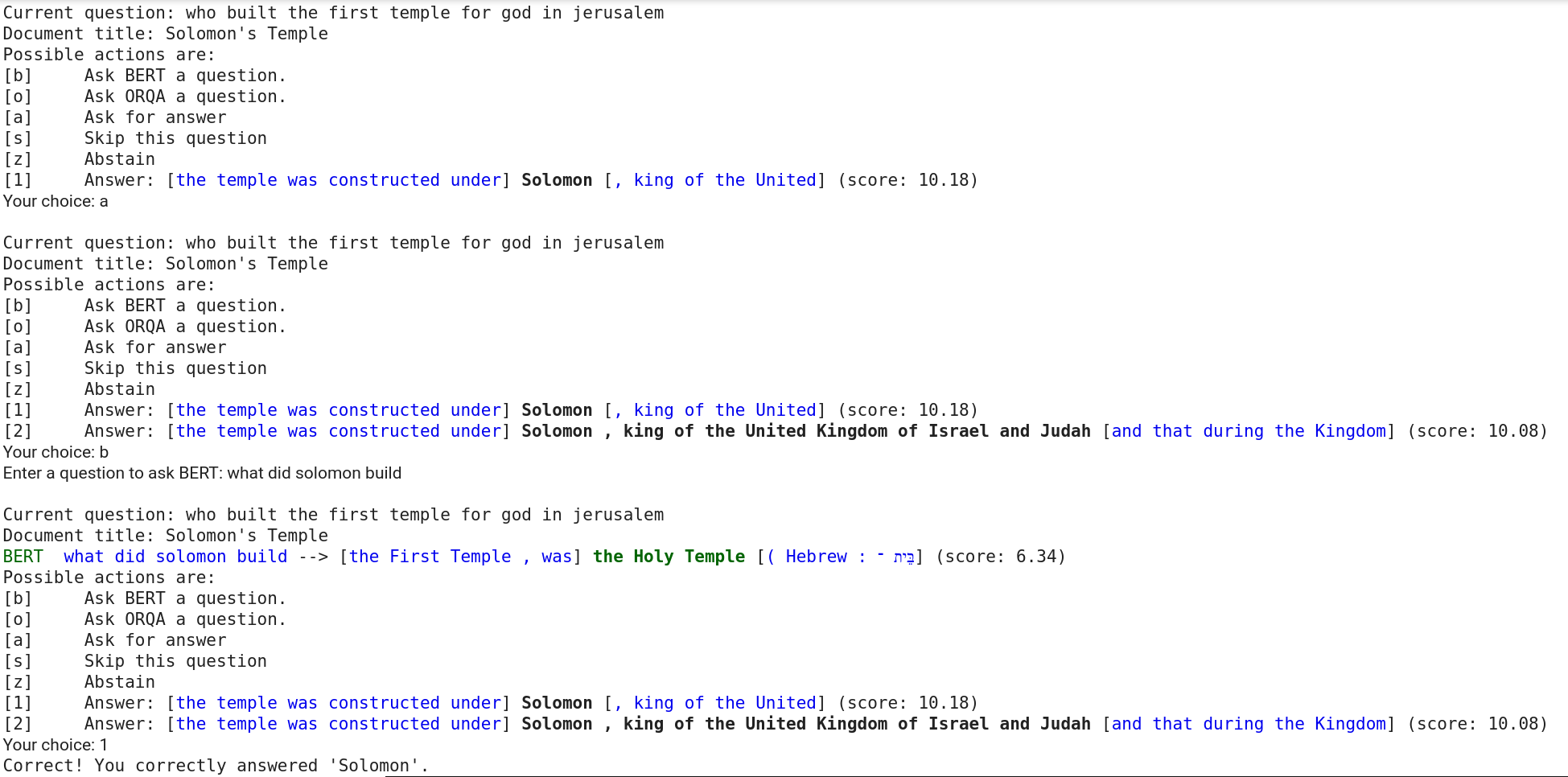}
\caption{Our interface for human \name{} experiments.  Users see a question from NQ, see candidates from base QA system, and find their meta answer.}
\label{fig:interface}
\end{figure*}

%\end{figure*}

\end{document}